\documentclass[runningheads]{llncs}
\usepackage{graphicx}
\usepackage{tikz}
\usepackage{comment}
\usepackage{amsmath,amssymb} % define this before the line numbering.
\usepackage{color}
\usepackage{multirow}
\usepackage[accsupp]{axessibility}  % Improves PDF readability for those with disabilities.
\usepackage{pifont}
\newcommand{\cmark}{\ding{51}}%
\newcommand{\xmark}{\ding{55}}%
\usepackage{booktabs}
\usepackage[pagebackref=true,breaklinks=true,colorlinks,bookmarks=false]{hyperref}
\usepackage[font=small,labelfont=bf]{caption}
\usepackage{subcaption}
\usepackage{cleveref}
\usepackage[width=122mm,left=12mm,paperwidth=146mm,height=193mm,top=12mm,paperheight=217mm]{geometry}

\newcommand{\revision}[1]{\textcolor[rgb]{0.0,0.0,0.0}{#1}}

\definecolor{newred}{HTML}{c0392b}
\definecolor{neworange}{HTML}{f39c12}
\definecolor{newblue}{HTML}{2980b9}
\newenvironment{packed_itemize}{
\begin{list}{\labelitemi}{\leftmargin=2em}
\vspace{-6pt}
 \setlength{\itemsep}{0pt}
 \setlength{\parskip}{0pt}
 \setlength{\parsep}{0pt}
}{\end{list}}

\begin{document}
% \linenumbers
\pagestyle{headings}
\mainmatter
\def\ECCVSubNumber{6199}  % Insert your submission number here

\title{\textcolor{newred}{M}\textcolor{neworange}{U}\textcolor{newblue}{GEN}: A Playground for Video-Audio-Text \textcolor{newred}{M}ultimodal \textcolor{neworange}{U}nderstanding and \textcolor{newblue}{GEN}eration} % Replace with your title

\begin{comment}
\titlerunning{ECCV-22 submission ID \ECCVSubNumber} 
\authorrunning{ECCV-22 submission ID \ECCVSubNumber} 
\author{Anonymous ECCV submission}
\institute{Paper ID \ECCVSubNumber}
\end{comment}
%******************

% \begin{comment}
\titlerunning{MUGEN}
\author{Thomas Hayes\inst{\star 1} \and
Songyang Zhang\thanks{equal contribution, ordered alphabetically}\inst{2} \and
Xi Yin\inst{1} \and Guan Pang\inst{1} \and Sasha Sheng\inst{1} \and Harry Yang\inst{1} \and Songwei Ge\inst{3} \and Qiyuan Hu\inst{1} \and Devi Parikh\inst{1}}
\authorrunning{T. Hayes et al.}
% \institute{Meta AI, Princeton NJ 08544, USA \and

\institute{\textsuperscript{1}Meta AI Research \hspace{0.2cm}
\textsuperscript{2}University of Rochester \hspace{0.2cm} \textsuperscript{3}University of Maryland\\
\email{\{thayes427,yinxi,gpang,sash,harryyang,isabellehu,dparikh\}@fb.com,\\szhang83@ur.rochester.edu,songweig@umd.com} \\ \textcolor{red}{https://mugen-org.github.io/}
}

% \end{comment}
%******************
\maketitle

\begin{abstract}
Multimodal video-audio-text understanding and generation can benefit from datasets that are narrow but rich. The narrowness allows bite-sized challenges that the research community can make progress on. The richness ensures we are making progress along the core challenges. To this end, we present a large-scale video-audio-text dataset MUGEN, collected using the open-sourced platform game CoinRun~\cite{cobbe2019quantifying}. We made substantial modifications to make the game richer by introducing audio and enabling new interactions. We trained RL agents with different objectives to navigate the game and interact with $13$ objects and characters. This allows us to automatically extract a large collection of diverse videos and associated audio. We sample $375$K video clips ($3.2$s each) and collect text descriptions from human annotators. Each video has additional annotations that are extracted automatically from the game engine, such as accurate semantic maps for each frame and templated textual descriptions. Altogether, MUGEN can help progress research in many tasks in multimodal understanding and generation. We benchmark representative approaches on tasks involving video-audio-text retrieval and generation. 
\revision{Our dataset and code are released at: \url{https://mugen-org.github.io/.}}  
%Both MUGEN and the enhanced game engine will be released to serve as a playground for multimodal research. 

\keywords{video, audio, language, multi-modal, retrieval, generation}
\end{abstract}

\begin{figure}[t]
    \centering
    \includegraphics[width=\textwidth]{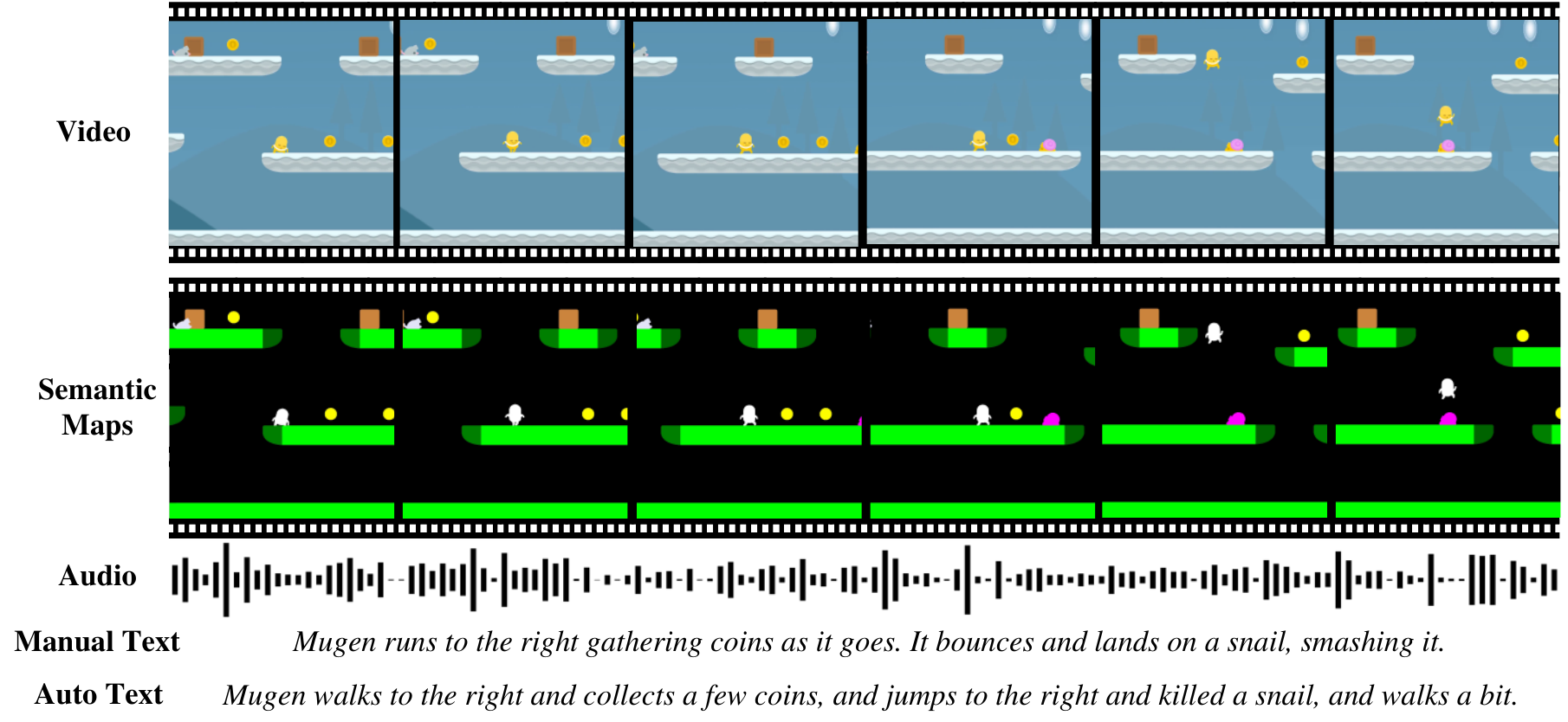}
    \caption{An example from our dataset. For each $3.2$s video clip, we have rich annotations including accurate semantic maps, synchronized audio, manual text collected from human annotators, and auto-text generated based on certain rules.} 
    \label{fig:illustration}
\end{figure}

%\vspace{-10pt}
\section{Introduction}
Research in multimodal understanding and generation brings together the subfields of vision and language in AI. Significant progress has been made on image-text understanding and generation tasks, such as CLIP~\cite{radford2021learning} for image-text retrieval and DALL-E~\cite{ramesh2021zero} for text-to-image generation. This progress has been made possible with large-scale image-text datasets~\cite{changpinyo2021conceptual,ordonez2011im2text,sharma2018conceptual,srinivasan2021wit,young2014image} that are collected from the web. However, progress in the video-text domain lags due to challenges in data collection and modeling of spatiotemporal information.

Many existing video-text datasets~\cite{xu2016msr,lei2018tvqa,zhou2018towards,miech2019howto100m} are collected in the wild and are proposed for understanding tasks such as video-text retrieval~\cite{miech2019howto100m}, video question answering~\cite{lei2018tvqa}, and generation tasks like video captioning~\cite{xu2016msr}. Yet the performance on these tasks is still far behind their image counterparts~\cite{chen2020uniter,li2020oscar,zhang2021vinvl} due to the challenges in understanding the complex dynamics in these in-the-wild videos. Moreover, such video-text pairs are too challenging for text-to-video generation, where more constrained datasets are used instead, e.g., bouncing MNIST~\cite{kahou2017ratm}, KTH~\cite{mittal2017sync}
and UCF-101~\cite{mazaheri2021video}. However, these are limited in actions and interactions between entities which are crucial to modeling real-world videos.

In this paper, we introduce MUGEN, a large-scale controllable video-audio-text dataset with rich annotations for multimodal understanding and generation. MUGEN is collected in a closed world based on the open-sourced platform game CoinRun~\cite{cobbe2019quantifying}. We have made substantial modifications to the game engine to make the videos more diverse (and delightful) by introducing audio, adjusting camera zoom and stabilization, and enabling new interactions between characters. We name the protagonist ``Mugen", and collected videos about Mugen's interactions with the other characters and objects.

To collect videos, we train reinforcement learning (RL) agents to navigate the world and record gameplay.
To increase video diversity and reduce bias towards the actions of any single agent, we trained $14$ RL agents with different objectives. 
We record $233$K videos of gameplay where the game environment is procedurally generated,
so there are no video duplicates.
We then sample $375$K $3.2$s video clips from this video set to collect text descriptions from human annotators (which we call ``manual text"). For each video clip, there are additional annotations that come for free: 1) audio is generated from a set of background music and foreground sound effects; 2) accurate semantic maps are generated for each frame using the game assets; 3) automatic text descriptions (``auto-text") are generated based on Mugen's actions and language templates. 
This results in $375$K video-audio-text samples in the MUGEN dataset. One example is shown in Figure~\ref{fig:illustration}.

\begin{table}[t]
\centering
   \caption{Comparison between MUGEN and other datasets.  Sent., Sem. and Cust. represent sentence, semantic and customizable. M, A and ASR represent descriptions that are manual annotated, automatically generated and translated from speech. R and G represent sound recorded with the video and generated based on the video.}
    \label{tab:dataset}
   \begin{tabular}{ccccccccc}
    \toprule
    \multirow{2}{*}{Dataset} & Video & Sent. & \multicolumn{3}{c}{Number of} & \multicolumn{3}{c}{Properties} \\
	\cmidrule(lr){4-6}
	\cmidrule(lr){7-9}
    & Content & Source & Sent. & Videos & Clips & Sem. & Audio & Cust. \\
    \midrule
    BMNIST~\cite{kahou2017ratm} & digit & A & - & - & - & \xmark & \xmark & \cmark \\
    KTH~\cite{mittal2017sync} & human action & A & 2K & - & 2K & \xmark & \xmark & \xmark \\
    TVR~\cite{lei2020tvr} & TV show & M & 11K & - & 22K & \xmark & R & \xmark \\
    %ANetCap~\cite{krishna2017dense} & human action & M & 100K & 20K & 100K & \xmark & R & \xmark \\
    %LSMDC~\cite{rohrbach2015dataset} & movie & script & 128K & 200 & 128K & \xmark & R & \xmark \\
    %DiDeMo~\cite{anne2017localizing} & open & M & 27K & 10K & 41K & \xmark & R & \xmark \\
    YouCook2~\cite{zhou2018towards} & cooking & M & 14K & 2K & 14K & \xmark & R & \xmark \\
    MSVD~\cite{chen2011collecting} & open & M & 70K & - & 2K & \xmark & R & \xmark \\
    %UCF-101~\cite{mazaheri2021video} & sports & M & ? & ? & ? & \xmark & R & \xmark \\
    %Kinetics~\cite{li2018video} & human action& M & ? & ? & ? & \xmark & R & \xmark \\
    A2D~\cite{xu2015can} & human action& M & 7K & 4K & 4K & \cmark & R\ & \xmark \\
    %VATEX~\cite{wang2019vatex} & human action & M & 825K & 41K & 41K & \xmark & R & \xmark \\
    %VIOLIN~\cite{liu2020violin} & TV show\&movie & M & 95K & 7K & 16K & \xmark & R & \xmark \\
    Charades~\cite{sigurdsson2016hollywood} & daily life & M & 16K & 10K & 10K & \xmark & R & \xmark \\
    FLINTSTONES~\cite{gupta2018imagine} & cartoon & M & 25K & - & 25K & \cmark & R & \xmark \\
    MSRVTT~\cite{xu2016msr} & human activity & M & 200K & 7180 & 10K & \xmark & R & \xmark \\
    %COIN~\cite{tang2019coin} & instructional & M & 46K & 12K & 46K & \xmark & R & \xmark \\
    %WebVid-2M~\cite{Bain21} & open & ASR & 2.5M & - & 2.5M & \xmark & R & \xmark \\
    HowTo100M~\cite{miech2019howto100m} & instructional & ASR & 136M & 1.2M & 136M & \xmark & R & \xmark \\
    \midrule
    MUGEN (ours) & platform game & M+A & 379K & 233K & 375K & \cmark & G & \cmark \\
    \toprule
    \end{tabular}
\end{table}

Table~\ref{tab:dataset} shows a comparison between MUGEN and other multimodal datasets. 
There are several advantages of MUGEN. 
First, the videos in MUGEN are collected in a closed world with a limited set of visually simple objects and scenes (i.e., simpler than in-the-wild datasets) but with diverse motions and interactions between entities that capture 
some of the core challenges in video understanding and generation (i.e., richer than other closed world datasets). Not only does the narrowness allow for bite-sized challenges to make progress on, it also alleviates the need for web-scale data and correspondingly massive compute resources in studying multimodal understanding and generation. 
Second, there are rich annotations for each video including accurate semantic maps, synchronized audio, and auto-text and manual text descriptions that can enable a wide variety of tasks in multimodal research.
Third, the game engine setup allows us to render videos at different resolutions on the fly, which is more flexible and storage efficient. Fourth, the game engine is modifiable, and once released, will allow the research community to collect more data to study a diverse set of problems.

MUGEN enables study of many multimodal video-audio-text tasks; in this paper, we focus on several tasks including retrieval and generation between all pairs of modalities. For the research community to make progress, it is vital to have consistent evaluation protocols. 
While many automatic metrics have been proposed for evaluation of generative models,
human judgement is still the gold standard.
Prior works have compared to ground-truth
for text~\cite{cui2018learning} or audio generation~\cite{chen2017deep}, but video generation evaluation is usually conducted by comparing to baselines because ground-truth is too challenging. This makes it difficult to compare methods and calibrate progress over time. Given that MUGEN is a closed world dataset with simplified visual elements, it is possible to compare to ground-truth videos for evaluation. In this paper, we conduct a comprehensive human evaluation for various cross-modal generation baselines.
We evaluate both the generation quality as well as faithfulness to input modality. We hope this evaluation protocol will be adopted by the community. We will make our evaluation interfaces publicly available.

We summarize the contributions of this paper as follows:
\begin{packed_itemize}
    \item We propose MUGEN, a large-scale dataset of $375$K video-audio-text samples with additional annotations of semantic maps and auto-text to facilitate research in multi-modal understanding and generation.
    \item We benchmark the performance of video-audio-text retrieval and generation between every pair of modalities in a unified framework.
    To our knowledge, this is the first work that benchmarks all these tasks on one dataset.
    \item We formulate a standard protocol for human evaluation of quality and faithfulness for four generation tasks.
    \item We will release the dataset and the game platform so the community can generate more data for a variety of tasks to push the field forward.
\end{packed_itemize}

\section{Related Work}

\vspace{1mm}
\noindent {\bf{Multimodal Datasets.}}
Existing multimodal datasets belong to two categories based on the visual content: open world (in-the-wild environments) and closed world (constrained environments). Open world datasets such as MSCOCO~\cite{lin2014microsoft}, ConceptualCaptions~\cite{changpinyo2021conceptual}, and WIT~\cite{srinivasan2021wit} are widely used for image-text research. CLEVR~\cite{johnson2017clevr} is a closed world dataset collected by arranging different 3D shapes on a clean background, which enables systematic progress in visual reasoning by reducing the complexity and bias from the real world.

Most video-text datasets are open world. MSRVTT~\cite{xu2016msr}, ANetCap~\cite{krishna2017dense}, MSVD~\cite{chen2011collecting}, and DiDeMo~\cite{anne2017localizing} contain videos of sports and human actions collected from the web. YouCook2~\cite{zhou2018towards} and HowTo100M~\cite{miech2019howto100m} contain instructional videos collected from YouTube. TVR~\cite{lei2020tvr}, TVQA~\cite{lei2018tvqa}, and LSMDC~\cite{rohrbach2015dataset} are collected from TV series and movies. Ego4D~\cite{grauman2021ego4d} is collected by people wearing an egocentric camera recording everyday activities around the world. Videos in these datasets contain complex backgrounds and diverse events, which makes them very challenging. Datasets from constrained environments, \textit{e.g.} Bouncing MNIST (BMNIST)~\cite{kahou2017ratm} and KTH~\cite{mittal2017sync},
have been proposed. These datasets don't capture some of the core challenges in videos
such as multiple entities interacting with each other in meaningful ways. 
FLINTSTONES~\cite{gupta2018imagine} is created from an animated series, but the scenes are too diverse for the size of the dataset.
In contrast, MUGEN simplifies the visual complexities of the scenes and objects, but captures complex motion and interactions between multiple entities. 

The text in existing datasets are either collected from humans~\cite{gupta2018imagine,xu2016msr,zhou2018towards} or extracted from  speech~\cite{miech2019howto100m}. Besides human descriptions, MUGEN also allows generating templated auto-text descriptions for videos of arbitrary lengths.

Most open world video-text datasets are associated with audio recorded from human speech and/or events.
AudioSet~\cite{gemmeke2017audio} and VGGSound~\cite{chen2020vggsound} are collected with video-audio pairs for audio event recognition. 
However, the video and audio are often not well-aligned. 
(E.g., the speech may describe things not related to or aligned with the video and background noise is common.) 
In contrast, the video and audio in MUGEN are synchronized based on Mugen's actions, 
making it feasible to study less explored tasks like audio generation from video or text.

\vspace{1mm}
\noindent {\bf{Multimodal Understanding and Generation.}}
Multimodal research typically involves four modalities: image, video, audio and text. Image-text tasks are widely studied, such as VQA~\cite{antol2015vqa}, image captioning~\cite{agrawal2019nocaps,johnson2016densecap,you2016image}, image-text retrieval~\cite{johnson2015image}, visual storytelling~\cite{huang2016visual}, text-to-image generation~\cite{reed2016generative}, etc.
Earlier methods aimed to design effective models for specific tasks~\cite{goyal2017making,li2019storygan,niu2021counterfactual,Yang_2019_ICCV}.
Later work~\cite{chen2020uniter,li2020oscar,zhou2020unified} focused on large-scale pre-training to learn cross-modal representations that can be transferred to various downstream tasks. More recently, CLIP~\cite{radford2021learning}, CogView~\cite{ding2021cogview}, and DALL-E~\cite{ramesh2021zero} leverage even larger-scale training to improve model generalization and zero-shot learning. FLAVA~\cite{singh2021flava} and Florence~\cite{yuan2021florence} were proposed as foundation models for both vision and language. 

Many video-text tasks have been proposed, such as video QA~\cite{lei2018tvqa}, video-text retrieval~\cite{xu2016msr}, video grounding~\cite{gao2017tall}, video captioning~\cite{xu2016msr}, text-to-video generation~\cite{gupta2018imagine}, etc.
Similar to image-text research, early approaches focused on a single task~\cite{gao2017video,gabeur2020mmt,le2020hierarchical,Zhang2021MS2DTAN}. Some recent work proposed novel architectures to learn task-agnostic video-text embeddings, such as MIL-NCE~\cite{miech19endtoend} and ClipBERT~\cite{lei2021less}. Compared to video-text retrieval and video captioning, text-to-video generation is relatively understudied, largely due to a lack of feasible datasets. Early methods~\cite{mittal2017sync,li2018video,liu2019cross} were evaluated on simple datasets like BMNIST~\cite{kahou2017ratm} and KTH~\cite{mittal2017sync}. 
Mazaheri and Shah~\cite{mazaheri2021video} annotated $10$ action classes from UCF-101~\cite{mazaheri2021video}. However, the limited motion in these datasets restricts the diversity of the collected text descriptions, making them sub-optimal for studying text-to-video generation.  

There are also efforts on audio, such as audio-text retrieval~\cite{koepke2022audio}, audio captioning~\cite{kim2019audiocaps}, audio-to-video generation~\cite{mama2021nwt}, video-to-audio generation~\cite{iashin2021taming}, etc. We explore video-audio-text retrieval and generation between all pairs of modalities on MUGEN,
and conduct extensive evaluations including human evaluation.

\section{MUGEN Dataset}
\begin{figure}[t]
\begin{subfigure}{.5\textwidth}
  \includegraphics[width=\linewidth]{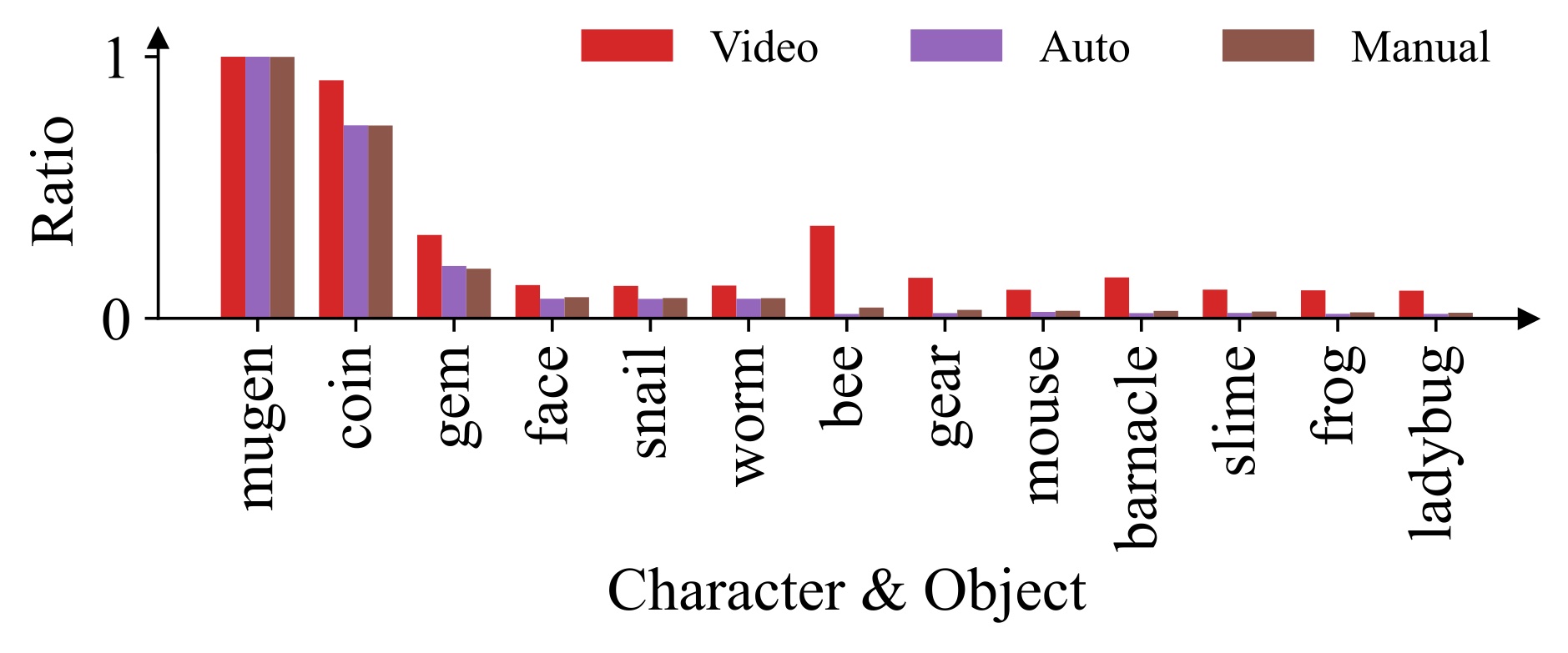}
  \caption{Characters and objects in video and text.}
  \label{fig:dist_auto}
\end{subfigure}
\begin{subfigure}{.5\textwidth}
\includegraphics[width=\linewidth]{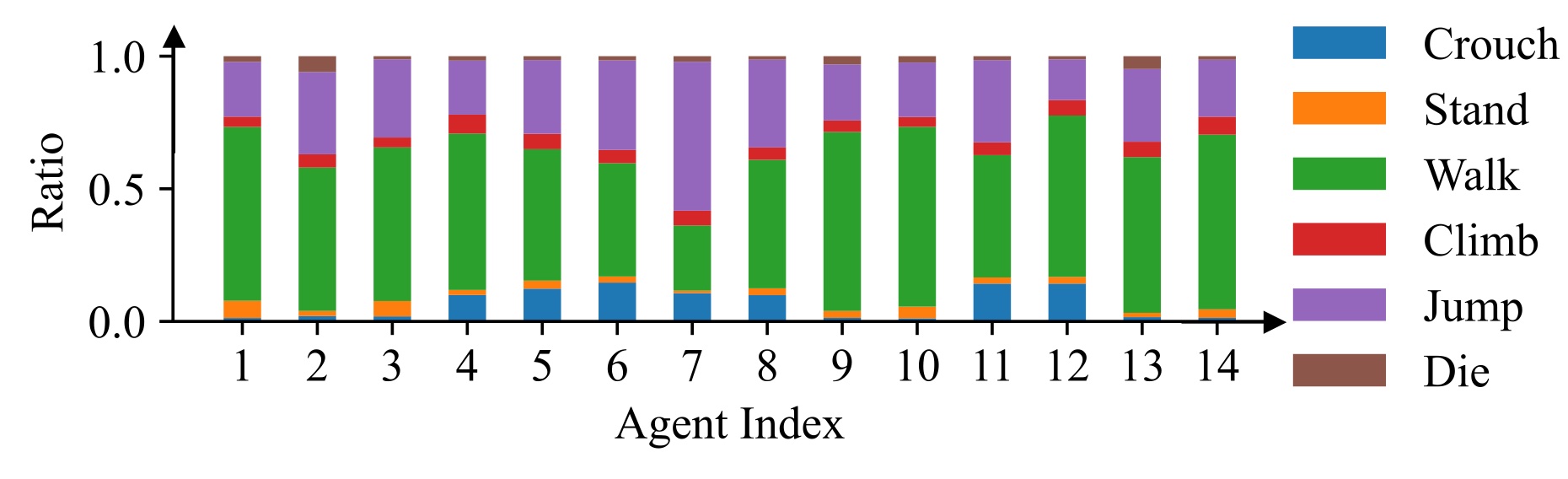}
 \caption{Mugen's poses across the 14 RL agents.}
 \label{fig:density_event}
\end{subfigure}
\begin{subfigure}{\textwidth}
\includegraphics[width=\linewidth]{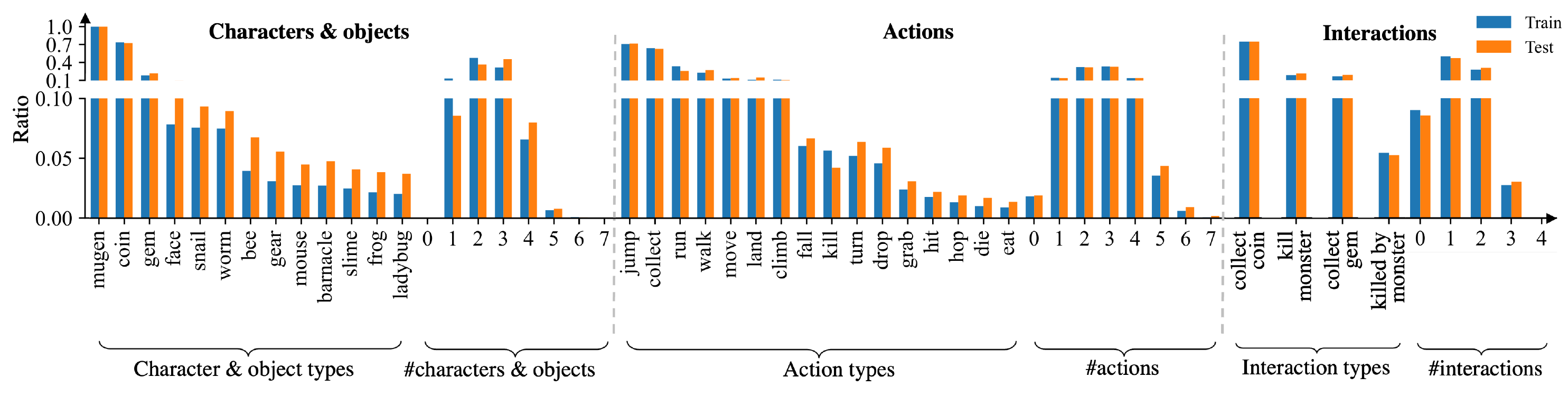}
  \caption{Characters, objects, actions and interactions in three splits and their occurrence per video.
  }
  \label{fig:dist_all_in_one}
\end{subfigure}
\caption{Distribution of characters, objects, actions and interactions.
}
\label{fig:distribution}
\end{figure}

\vspace{1mm}
\noindent {\bf{Environment.}}
In-the-wild video understanding and generation poses several challenges, including understanding motion of objects, interactions among objects, physics, camera vs. object and scene motion, 3D depth of scenes, diverse object appearances and semantics, lighting conditions, etc. Our goal was to develop a dataset that is rich along some of these dimensions, but narrower along others, to enable focused advances in some of the core challenges of multimodal video research. Specifically, we desired a closed world dataset where physics are simplified, the camera angle is fixed, the number of objects is limited, and lighting is consistent. Yet, we sought diverse motions and interactions between entities (dataset statistics are shown in Figure~\ref{fig:distribution}).
For these purposes, we chose an open source video game which (1) enables training RL agents to collect video data at scale and (2) gives access to the game engine that provides additional high quality annotations for free, such as precise frame-level semantic maps and automatic text descriptions. Amongst open source games, we chose OpenAI's CoinRun~\cite{cobbe2019quantifying} because of its ease of modification.

OpenAI's CoinRun is a platform game developed for quantifying generalization of RL agents~\cite{cobbe2019quantifying,edwards2019imitating,igl2019generalization}. The game has a single main character (who we call Mugen) with the objective to collect coins without being killed by monsters. Each level has a number of coins and monsters, and the level ends when Mugen collects all coins, Mugen is killed by a monster, or the level times out after $21$ seconds. The environment is procedurally generated, with each level having a unique configuration of platforms, coins, and monsters. 

We made a number of modifications to increase the diversity of game events and enhance richness, such as adding audio, slowing game physics, adjusting camera zoom and stabilization, and enabling new interactions between characters. Altogether, our updated version of CoinRun features Mugen, $10$ monsters, coin and gem objects, and $2$ world themes, space and snow. Mugen can take $16$ different actions (see Figure~\ref{fig:dist_all_in_one} for the most frequent actions). Monsters differ in their action vocabulary; some walk, others hop, and one flies. A full list of modifications, before and after videos highlighting these changes, and images of these characters, objects, and themes can be found in the appendix.  

\vspace{1mm}
\noindent {\bf{Audio.}} The audio consists of two layers, sound effects and background music. We chose $8$ sound effects corresponding to Mugen's core actions:
walk, jump, collect coin, kill monster, power-up, climb ladder, bump head, die.
Each sound effect is triggered by these actions, and one sound effect plays at a time. Background music features $2$ themes for the space and snow game themes. Background music is layered with the sound effect audio to produce the full audio track.

\vspace{1mm}
\noindent {\bf{Video Collection.}}
We train RL agents to navigate the environment and collect gameplay videos. We use an IMPALA-CNN architecture~\cite{espeholt2018impala} and train agents with Proximal Policy Optimization~\cite{schulman2017proximal}. Inputs to the agent include the current game frame and the agent’s velocity. The agent's performance in the game is immaterial to us; 
we care about maximizing the diversity of video data. To this end, we trained $14$ agents with modified reward functions to achieve different behaviors.
For example, decreasing the reward discount factor makes the agent more myopic and risk-tolerant, so the agent dies frequently. Figure~\ref{fig:density_event} shows the distribution of Mugen's poses for each agent where the variation in time spent in different poses indicates differing actions.
To further increase diversity, we ensured that the seed for map procedural generation is always unique. We have verified there are no duplicate videos in MUGEN.

To efficiently handle large-scale game video data and enable easy data customization (e.g., swapping characters or objects, changing background), we do not save the rendered videos. 
Instead, we save all metadata such as world layout and character movements in a json format, from which we can render RGB frames and pixel-accurate segmentation maps at any resolution up to $1400\times1400$ on-the-fly, resulting in more efficient data storage.~\footnote{Storage is $>100\times$ smaller than $1024\times1024$ videos stored with lossless encoding.} 

We recorded $233$K videos of gameplay ranging from $3.2$s to $21$s (level timeout) at $30$ frames per second. Each video corresponds to a whole level of gameplay. We will release the game engine so others can customize the data environment or agents for their own purposes.

\vspace{1mm}
\noindent {\bf{Manual Text.}}
We split the $233$K videos into $3.2$s (96 frames) clips and ask annotators to describe in 1-2 sentences what happens in the short video.
After filtering low quality annotations,  
MUGEN consists of $378,902$ text descriptions for $375,368$ video clips.~\footnote{A very small portion of the clips have more than one description.} Refer to the appendix for the annotation interface and details on annotation quality control.

\vspace{1mm}
\noindent {\bf{Auto-Text.}}
In addition to collecting human annotation, we also developed a template-based algorithm to automatically generate textual descriptions for videos based on game engine metadata. See the appendix for details.

Note that both video and auto-text can be generated automatically. We can generate arbitrary amounts of video-text data with arbitrary lengths. This makes it feasible to study more tasks where manual annotations are expensive to acquire, such as text-conditioned long video generation~\cite{ge2022long}, video grounding~\cite{zeng2020dense}, and dense video captioning~\cite{krishna2017dense}.
Auto-text is also highly structured in nature. This simplifies the text
and improves model explainability since each action and interaction in the video has a unique description in the text.

\vspace{1mm}
\noindent {\bf{Dataset Statistics.}}
In total, MUGEN consists of $375$K $3.2$s video clips paired with $379$K manual text descriptions, as well as $233$K longer ($3.2$s to $21$s) videos. Each video clip or long video also comes with semantic maps, auto-text, and audio. 
There are $11$ characters, $2$ objects, $16$ different actions for Mugen, and $4$ classes of interactions with other objects and characters: collect coin, collect gem (power up), kill monster, killed by monster.

We first analyze the occurrences of characters and objects in video, manual text, and auto-text in Figure~\ref{fig:dist_auto}.~\footnote{The occurrence of one character is counted at most once in each video/text.} We observe that not all characters and objects appearing in the video are mentioned in the text. This is because annotators are more likely to describe characters that interact with Mugen than those in the background. Given the unbalanced distribution of characters and objects, when splitting our dataset, we sample fewer videos featuring only Mugen or Mugen's interaction with coins for the validation and test sets.
Both validation and test sets contain only one manual text per video. This results in $349,666$, $12,851$, $12,851$ video clips in training, validation, and test sets, respectively.

The distributions of characters and objects, actions, and interactions 
are shown in Figure~\ref{fig:dist_all_in_one}. ``Jump'' and ``collect'' are the top $2$ most frequent actions, consistent with ``collect coin'' being the most frequent class of interaction (this is CoinRun after all!). The rarest interaction type is Mugen being killed by a monster.
We also show the distribution of the number of characters and objects, actions, and interactions in each video. 
Most videos contain 2-4 characters and objects, 2-4 actions, and 2-3 interactions.
This is more diverse than other closed world datasets with one or two digits moving~\cite{kahou2017ratm} or a single person moving in a scene~\cite{mittal2017sync}. 
We also show the location heatmap of each charchater/object and temporal heatmap of each action/interaction in the appendix.

As shown in Table~\ref{tab:dataset}, MUGEN is several orders of magnitude larger than existing closed world datasets such as BMNIST~\cite{kahou2017ratm}, KTH~\cite{mittal2017sync} and FLINTSTONES~\cite{gupta2018imagine}. 
While it is smaller than some open world datasets including HowTo100M~\cite{miech2019howto100m} and WebVid-2M~\cite{Bain21}, it is also visually less diverse, making it feasible to train effective models without having to work with web-scale data. Moreover, our dataset provides audio aligned with video, accurate frame-level semantic maps, and automatically generated text descriptions which enable studying a variety of tasks. 
Finally, this dataset is customizable with the released game engine, so the community can generate more data of different distributions.

\section{Video-Audio-Text Retrieval and Generation}

While MUGEN can enable many tasks, we focus on retrieval and generation between every pair of modalities. We first present the cross-modal retrieval framework and then a unified pipeline for cross-modal generation.

\subsection{Video-Audio-Text Retrieval.}
Cross-modal retrieval, which retrieves samples from one modality given a query from another, is a fundamental task with many real-world applications. For example, text-to-video retrieval is widely used for video search. 

We use an encoder $F_{x}$ to map input $x$ of each modality to a feature vector $\mathbf{f}_x={F}_{x}(x)$. It is projected into a joint embedding space $\mathbf{e}_x=\mathbf{f}_x\cdot \mathbf{W}_x$, where $\mathbf{W}_x$ are the learnable parameters.
Given inputs $p$ and $q$ from two modalities P and Q, the similarity can be computed by a scaled cosine function,  $s(p,q)=cos(\mathbf{e}_{p}, \mathbf{e}_{q})\cdot e^{\tau_{PQ}}$, where $\tau_{PQ}$ is a learnable temperature parameter. 
The matching loss $L_{PQ}$ is computed as:
\begin{equation}
    L_{PQ}=-\frac{1}{2N}\sum_{i=1}^N(\log(\frac{e^{s{(p_i,q_i)}}}{\sum_{j=1}^Ne^{s{(p_i,q_j)}}})+\log(\frac{e^{s{(p_i,q_i)}}}{\sum_{k=1}^Ne^{s{(p_k,q_i)}}})),
\end{equation}
where $N$ is the number of samples in a batch, $p_i$ and $q_i$ represent the $i$th sample from $P$ and $Q$ modalities within the batch.

We train three models with $L_{VA}$, $L_{VT}$ and $L_{AT}$ separately for video(V)-audio(A), video-text(T), and audio-text retrieval.
For comparison, we also sum three losses to learn a joint model. 

During inference, to retrieve samples from modality $P$ given a query from modality $Q$, we rank the samples based on the similarities $s(p,q)$. 
To retrieve modality $P$ based on queries from two modalities $Q$ and $R$, we sum the similarity from two modalities,  $s(p,q)+s(p,r)$.
$s(p,q)$ and $s(p,r)$ can either come from two models independently trained by $L_{PQ}$ and $L_{PR}$ or the joint model.

\begin{figure}[t]
\centering
\includegraphics[width=\textwidth]{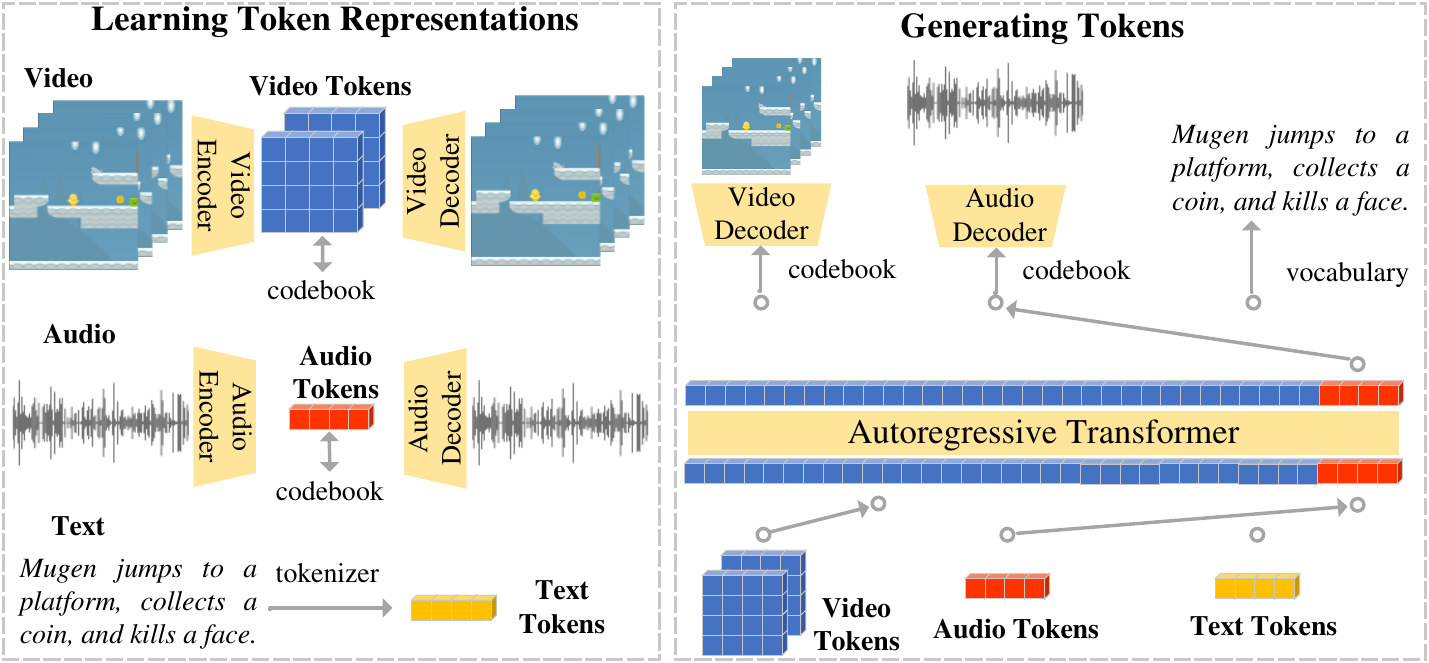}
\caption{A unified framework for generation between every pairs of modalities. The right part shows an example of video-to-audio generation.
}
\label{fig:videogpt}
\end{figure}

\subsection{Video-Audio-Text Generation}
Cross-modal generation has gained increasing interest in recent years. Amongst video-audio-text cross-modal generation tasks, video-to-text generation (video captioning) is most studied, while other tasks (video-to-audio, text-to-video, etc.) are relatively under-explored. 

Inspired by the success of using a VQ-VAE~\cite{van2017neural} and transformer for image~\cite{ramesh2021zero} and video~\cite{yan2021videogpt} generation, we adopt a similar and unified framework for cross-modal generation, as shown in Figure~\ref{fig:videogpt}.
For each modality, we first learn a discrete codebook to encode the data. Then an decoder-only transformer is used to learn token generation from one modality to another.

\vspace{1mm}
\noindent {\bf{Learning Token Representations.}}
For video representation, we train a 3D VQ-VAE to learn a codebook following the training losses in~\cite{yan2021videogpt}. The encoder is used to encode videos as inputs for the transformer during training, and the decoder is used for video generation during inference.
Similarly, we train a 1D VQ-VAE to learn audio compression following the training losses in~\cite{dhariwal2020jukebox}. 
For text representation, we learn a tokenizer from manual text in the training set.  

\vspace{1mm}
\noindent {\bf{Generating Tokens.}}
We use a decoder-only transformer to learn auto-regressive token generation. During training, the input to the transformer is a sequence of tokens concatenated from modality $P$ and $Q$. Video tokens are flattened from 3D latent codes into 1D. Text tokens are truncated or padded to have the same length. Causal attention is used where each token can only attend to prior tokens. The transformer learns to predict the token ids at every location. The loss functions for the two modalities are summed as the training objective, similar to DALL-E~\cite{ramesh2021zero}. 
During inference, given the tokens from $P$, we auto-regressively generate all tokens for $Q$. For audio or video, we use the predicted tokens to look up the codebook embeddings and feed them into VQ-VAE's decoder to reconstruct the video or audio. For text, the vocabulary is used to reconstruct the sentences.

\section{Experiments}
\subsection{Video-Audio-Text Retrieval}

\begin{table}[t]
    \centering
    \caption{Performance comparison on video(V)-audio(A)-text(T) retrieval. For retrieval in modality $P$, $Q+R$ denotes the ensemble of two models independently trained by $L_{PQ}$ and $L_{PR}$; $Q+R*$ denotes the joint model. Recalls are shown in percentage($\%$).}
    \begin{tabular}{cccc}
    \toprule
    Query & \multicolumn{3}{c}{Video Retrieval} \\
    \cmidrule(lr){2-4}
    Type & R1 & R5 & R10 \\
     \midrule
    A & $58.59$ & $88.83$ & $94.41$ \\
    T & $8.54$ & $22.50$ & $31.71$ \\
    A+T & $\bf 81.50$ & $\bf 96.10$ & $\bf 98.26$ \\
    A+T* & $62.54$ & $87.33$ & $92.62$ \\
    \bottomrule
    \end{tabular}
    \hfill
    \begin{tabular}{cccc}
    \toprule
    Query & \multicolumn{3}{c}{Audio Retrieval} \\
    \cmidrule(lr){2-4}
    Type & R1 & R5 & R10 \\
     \midrule
    V & $61.14$ & $88.99$ & $94.59$ \\
    T & $2.40$ & $8.35$ & $13.38$ \\
    V+T & $\bf 69.59$ & $\bf 92.48$ & $\bf 96.42$ \\
    V+T* & $41.83$ & $73.04$ & $82.68$ \\
    \bottomrule
    \end{tabular}
    \hfill
    \begin{tabular}{cccc}
    \toprule
    Query & \multicolumn{3}{c}{Text Retrieval} \\
    \cmidrule(lr){2-4}
    Type & R1 & R5 & R10 \\
     \midrule
    V & $10.61$ & $25.72$ & $34.70$ \\
    A & $2.95$ & $9.36$ & $14.80$ \\
    V+A & $\bf 11.68$ & $\bf 27.13$ & $\bf 36.60$ \\
    V+A* & $10.95$ & $26.24$ & $35.33$ \\
    \bottomrule
    \end{tabular}
    \label{tab:retrieval_peformance}
\end{table}

The retrieval task is to find the true match from the test set in one modality, given queries of either one of the other modalities or both modalities. For retrieval based on queries from two modalities, we compare the ensemble of two separately trained models and the joint model.
We report recall at rank $1$, $5$ and $10$. 
For all experiments, if not specified, the video dimension is $256$$\times$$256$$\times$$32$, where the $32$ frames are evenly sampled from the $96$ frames to save computation.

\vspace{1mm}
\noindent {\bf{Implementation Details.}}
Pre-trained models are used as initialization including ResNet-$18$~\cite{he2016deep} pre-trained on VGGSound~\cite{chen2020vggsound} for the audio encoder, S3D~\cite{xie2018rethinking} pre-trained on Kinetics $400$~\cite{kay2017kinetics} for the video encoder, and DistilBERT~\cite{sanh2019distilbert} for the text encoder.
The parameters in the text encoder are fixed~\footnote{Our initial experiments show unstable training with learnable text encoder.} and the other two encoders are learnable. 
The temperature $\tau_{PQ}$ is initialized as $0.07$ and the maximum is $100$, the learning rate is $0.001$, and batch size is $16$. All models are trained for $400$K steps and checkpoints are selected based on the validation set.

\vspace{1mm}
\noindent {\bf{Results.}}
The results are shown in Table~\ref{tab:retrieval_peformance}. We have the following observations: 1) across all single modality retrieval tasks, video-to-audio and audio-to-video retrieval perform the best
and text-to-audio and audio-to-text perform the worst. This is because video and audio are synchronized, while audio and text are only sparsely aligned on Mugen's actions and interactions; different text descriptions can map to similar audio samples. 
2) retrieval based on two modalities with an ensemble of models ($P+Q$) consistently outperforms single modality retrieval. This is because the other modality can provide complementary information. For example, text contains information of Mugen's moving direction that is available in video but not audio. 
3) the performance of the joint model ($P+Q^*$) typically falls between the separately trained models,
which indicates that it is challenging to learn a joint embedding space for all modalities.

\subsection{Video-Audio-Text Generation}
We evaluate cross-modal generation for all pairs of modalities. We use V, A, T to denote video, audio, text, and P2Q to represent the task (e.g., T2V for text-to-video generation).
We focus on quantitative evaluations of the quality of the output and faithfulness to the input. 
%Qualitative results are in supp. mat. 

\begin{table}[!h]
\centering
\caption{Performance comparison on all generation tasks. D.S. denotes the down-sampled training set. F(V/A)D denotes FVD for video quality and FAD for audio quality. R.Sim. denotes the Relative Similarity to evaluate the faithfulness to the input. ``B4'', ``M.'', ``R.'' and ``C.'' denote BLEU4, METEOR, ROUGE and CIDEr. ``Q." and ``F." stand for quality and faithfulness. Video frame lengths map to different frame rates (8/16/32 represent 2.5/5/10 frames per second). Audio token lengths map to different compression ratios (68/137/275/551 represent 1024/512/256/128 compression ratios in VQ-VAE). All metrics except F(V/A)D are shown in percentage (\%). All metrics are better when higher except F(V/A)D, which is the lower the better.}
\begin{tabular}{cccccccccccccc}
\toprule
Out & In & Train & Out & In & Text & \multicolumn{6}{c}{Auto} & \multicolumn{2}{c}{Human}\\
\cmidrule(lr){7-12}
\cmidrule(lr){13-14}
Mod. & Mod. & Data & Len. & Len. & Type & F(V/A)D & R.Sim. & B4 & M. & R. & C. & Q. & F. \\
\midrule
\multirow{7}{*}{Text} & \multirow{3}{*}{Video} & \multirow{3}{*}{Full} & \multirow{3}{*}{-} & $8$ & \multirow{3}{*}{M} & \multirow{3}{*}{-} & $83.5$ & $7.4$ & $20.2$ & $27.9$ & $18.2$ & - & - \\
& & & & $16$ & & & $101.5$ & $\bf 7.8$ & $20.8$ & $28.7$ & $\bf 20.2$ & - & - \\
& & & & $32$ & & & $\bf 108.0$ & $\bf 7.8$ & $\bf 21.3$ & $\bf 29.1$ & $19.9$ & $\bf 31.3$ & $\bf 42.6$ \\
\cmidrule(lr){2-14}
& \multirow{4}{*}{Audio} & \multirow{4}{*}{Full} & \multirow{4}{*}{-} & $68$ & \multirow{4}{*}{M} & \multirow{4}{*}{-} & $101.0$ & $6.0$ & $19.3$ & $26.5$ & $14.1$ & \multirow{4}{*}{-} & \multirow{4}{*}{-} \\
& & & & $137$ & & & $103.9$ & $6.3$ & $19.3$ & $26.6$ & $14.4$ & & \\
& & & & $275$ & & & $106.7$ & $6.5$ & $19.3$ & $26.8$ & $14.7$ & & \\
& & & & $551$ & & & $\bf 107.5$ & $\bf 6.7$ & $\bf 19.4$ & $\bf 27.1$ & $\bf 15.5$  & & \\
\midrule \midrule
\multirow{14}{*}{Video} & \multirow{10}{*}{Text} & \multirow{5}{*}{Full} & $8$ & \multirow{5}{*}{-} & M & $112.7_{\pm0.2}$ & $39.5$ & $5.1$ & $15.2$ & $21.7$ & $11.1$ & - & - \\
& & & $16$ & & M & $72.7_{\pm2.0}$ & $63.9$ & $7.3$ & $18.5$ & $26.5$ & $15.3$ & - & - \\
& & & $32$ & & M & $\bf 61.0_{\pm0.6}$ & $64.9$ & $8.1$ & $19.9$ & $28.1$ & $\bf 19.2$ & $\bf 17.0$ & $\bf 31.6$ \\
& & & $32$ & & A & $140.7_{\pm3.1}$ & $14.3$ & $6.4$ & $17.8$ & $25.3$ & $14.9$ & $9.2$ & $11.7$ \\
& & & $32$ & & M+A & $61.4_{\pm1.0}$ & $\bf 66.5$ & $\bf 8.2$ & $\bf 20.0$ & $\bf 28.2$ & $19.1$ & - & - \\
\cmidrule(lr){3-14}
& & \multirow{5}{*}{D.S.} & $8$ & \multirow{5}{*}{-} & M & $112.7_{\pm1.1}$ & $42.0$ & $5.0$ & $15.3$ & $21.9$ & $11.0$ & - & - \\
& & & $16$ & & M & $72.2_{\pm1.7}$ & $69.5$ & $7.3$ & $18.6$ & $26.7$ & $15.8$ & - & - \\
& & & $32$ & & M & $62.0_{\pm0.7}$ & $70.6$ & $7.9$ & $20.0$ & $28.2$ & $19.0$ & $\bf 18.8$ & $\bf 35.7$ \\
& & & $32$ & & A & $151.7_{\pm3.4}$ & $20.5$ & $6.2$ & $17.7$ & $24.9$ & $14.2$ & $12.1$ & $13.1$ \\
& & & $32$ & & M+A & $\bf 61.0_{\pm1.4}$ & $\bf 72.2$ & $\bf 8.2$ & $\bf 20.2$ & $\bf 28.4$ & $\bf 19.7$ & - & - \\
\cmidrule(lr){2-14}
& \multirow{5}{*}{Audio} &  \multirow{5}{*}{Full} & \multirow{5}{*}{$32$} & $68$ & \multirow{4}{*}{-}  & $64.0_{\pm0.9}$ & $79.9$ & \multirow{4}{*}{-} & \multirow{4}{*}{-} & \multirow{4}{*}{-} & \multirow{4}{*}{-} & - & - \\
& & & & $137$ & & $66.4_{\pm0.4}$ & $91.4$ & & & & & - & - \\
& & & & $275$ & & $\bf 63.4_{\pm0.4}$ & $93.1$ & & & & & $\bf 17.6$ & $\bf 37.1$ \\
& & & & $551$ & & $64.5_{\pm1.0}$ & $\bf 93.6$ & & & & & -  & - \\
%& & & & $551$ & & $29.6_{\pm2.1}$ & $19.6$ & & & & & $\bf 18.36$  & $33.79$ \\
\midrule \midrule
\multirow{8}{*}{Audio} & \multirow{4}{*}{Video} & \multirow{4}{*}{Full} & $68$ & \multirow{4}{*}{$32$} & \multirow{4}{*}{-} & $523.8_{\pm1.0}$ & $86.5$ & \multirow{4}{*}{-} & \multirow{4}{*}{-} & \multirow{4}{*}{-}  & \multirow{4}{*}{-}  & - & - \\
& & & $137$ & & & $128.5_{\pm0.4}$ & $95.4$ & & & & & - & - \\
%& & & $137$ & & & $88.6_{\pm0.3}$ & $26.7$ & & & & & $16.21$ & $\bf 11.72$ \\ # with human results
& & & $275$ & & & $52.3_{\pm0.5}$ & $\bf 96.7$ & & & & & $\bf 15.2$ & $\bf 31.1$ \\
& & & $551$ & & & $\bf 50.0_{\pm0.8}$ & $92.7$ & & & & & - & - \\
%& & & $551$ & & & $46.7_{\pm0.5}$ & $17.7$ & & & & & $\bf 21.48$ & $9.77$ \\ # with human results
\cmidrule(lr){2-14}
& \multirow{4}{*}{Text} & \multirow{4}{*}{Full} & $68$ & \multirow{4}{*}{-} & \multirow{4}{*}{M} & $574.0_{\pm2.8}$ & $\bf 91.3$ & $7.0$ & $\bf 18.5$ & $26.5$ & $15.3$ & \multirow{4}{*}{-} & \multirow{4}{*}{-} \\
& & & $137$ & & & $171.1_{\pm2.2}$ & $88.8$ & $6.9$ & $\bf 18.5$ & $26.5$ & $15.0$ & & \\
& & & $275$ & & & $\bf 93.7_{\pm1.6}$ & $86.9$ & $\bf 7.1$ & $\bf 18.5$ & $\bf 26.9$ & $\bf 16.2$ & & \\
& & & $551$ & & & $109.4_{\pm2.3}$ & $78.7$ & $6.9$ & $18.1$ & $26.1$ & $15.9$ & & \\
\bottomrule
\end{tabular}
\label{tab:gen_comp}
\end{table}

\vspace{1mm}
\noindent {\bf{Implementation Details.}}
The 3D VQ-VAE is similar to~\cite{yan2021videogpt} except that we use a kernel size of $3$, which significantly sped up training compared to the original kernel size of $4$. We use a down-sample ratio of $32$$\times$$32$$\times$$4$ for video compression \revision{and a vocabulary of size $2048$}. The 3D VQ-VAE is trained for $600$K steps with a learning rate of $0.003$ and a batch size of $8$. The 1D VQ-VAE for audio features non-causal, dilated 1D convolutions where the dilation is grown by a factor of $3$. \revision{The vocabulary size is $1024$.} Audio sample rate is $22$kHz. The 1D VQ-VAE is trained for $1$M steps with a learning rate of $0.0003$ and a batch size of $4$. We use Byte-Pair Encoding (BPE)~\cite{sennrich-etal-2016-neural,gage1994new} for text tokenization and train a tokenizer from the manual text annotations in the training set.

All P2Q generation models are trained with the same transformer architecture and optimization hyper-parameters. For the transformer, we use $12$ layers with a hidden dimension of $768$ and $8$ attention heads. All models are trained for $600$K steps with a learning rate of $0.0003$ and batch size of $4$. Model checkpoints are selected based on the performance on the validation set.

\vspace{1mm}
\noindent {\bf{Inference.}}
During inference, we perform token sampling from the estimated distribution with filtering. For video or audio generation, we use top-k$=100$ and top-p$=0.9$ for filtering. For text generation, we use top-k$=1$, which is the same as beam search~\cite{anderson2016guided} with size $1$ in the common captioning setup. 

\vspace{1mm}
\noindent {\bf{Automatic Evaluation.}}
For text generation, we use metrics that are widely used in captioning evaluation including BLEU4, METEOR, ROUGE, and CIDEr. For video quality, we follow prior practices and use I3D pre-trained on Kinetics 400 to calculate FVD. For audio quality, we use the pre-trained audio encoder on VGGSound to calculate FAD. 
To automatically evaluate faithfulness to input, we propose a new metric Relative Similarity (R.Sim.) that leverages the retrieval models. Specifically, we calculate the average similarity between the input and output divided by the average similarity between the input and the ground truth.
For T2V and T2A generation, we use the V2T and A2T models applied on the generated video/audio to calculate the captioning metrics. 

\vspace{1mm}
\noindent {\bf{Human Evaluation.}}
We establish a human evaluation protocol to calibrate towards the Ground Truth (GT). We randomly selected $512$ samples from the test set and manually inspected the descriptions to ensure the samples were diverse and not too simple (e.g., to avoid multiple samples where Mugen simply jumps onto a platform).
For each task, we evaluate both quality and faithfulness to the input. As it is not straightforward for humans to judge the alignment between audio and text, we do not evaluate T2A and A2T but focus on the other cross-modal generation tasks. For quality, we ask human judges to select the higher fidelity sample (video, audio, or text) between the generation and GT. For faithfulness, human judges are asked to select the media which better aligns with the input~\footnote{We will release the annotation UIs for others to follow this protocol.}. Each comparison is evaluated by $5$ judges and the majority vote is taken. We report the percentage of samples that are chosen over the GT as the final metric. The upper bound for these evaluations is around $50\%$ when a human judge cannot tell the difference between the generation and the GT.

We remove confounding factors in the comparisons. For instance, for video comparison, we render the ground-truth video using the same theme (snow or space), frame rate, and resolution as the generated video. For generated text, several post-processing steps are used: capitalize the first letter of the first word in each sentence, use ``Mugen" to replace ``mugen" in all places, and remove duplicated spaces.  We use Amazon Mechanical Turk for human evaluation. 

\vspace{1mm}
\noindent {\bf{Text Generation from Video or Audio.}}
As shown in Table~\ref{tab:gen_comp}, we vary the video frame rate and audio compression ratio (a higher compression ratio results in fewer tokens) for comparison. For V2T generation, higher frame rate leads to stronger performance. Human evaluation shows high faithfulness with $42.6\%$ of samples chosen over GT, and relatively lower quality with $31.25\%$ of samples considered more realistic than GT. For A2T generation, a smaller compression ratio (more tokens) is better. A2T performs worse than V2T as video-text are more densely aligned than audio-text. 

\vspace{1mm}
\noindent {\bf{Video Generation from Audio or Text.}} 
For T2V generation, we experiment with training using manual text and auto-text.
As mentioned earlier, we balanced the characters in the validation and test sets. Correspondingly, to study the effects of data balancing, we also generate a smaller training set with $233$K samples by down-sampling videos with Mugen or Mugen and coins only. 
As shown in Table~\ref{tab:gen_comp}, for T2V generation, we have the following observations: 
1) Larger frame rate leads to better performance in all automatic metrics. 2) Auto-text performs worse than manual text 
and cannot noticeably improve performance when combined with manual text. This is because we evaluate on manual text for all comparisons.
We hypothesize that auto-text may be useful when manual text is not available or is limited. 3) Models trained on the down-sampled training set consistently outperform those on the full set. Future work can explore other sampling strategies to fully utilize the training set. 4) Human evaluation results show better faithfulness compared to quality. The trends between automatic metrics and human evaluation results are similar. 

For A2V generation, a smaller compression ratio (longer token sequence) leads to better quality and faithfulness in the automatic metrics. %with the optimal achieved at $275$ audio token length. 
Human evaluation shows higher faithfulness compared to quality, similar to the T2V task.

\vspace{1mm}
\noindent {\bf{Audio Generation from Video or Text.}}
\revision{For audio generation, generating a longer audio sequence (less compression) leads to better quality in FAD for both T2A and V2A. But the R.Sim. may not follow the same trend. Human evaluation also shows higher faithfulness than quality, similar to other tasks.}
%As for audio generation, similarly, we observe that generating a longer audio sequence (less compression) leads to better performance for both T2A and V2A. The best performance is achieved with audio token length $275$ for V2A and $551$ for T2A (except R.Sim. score).

When comparing the human evaluation results for all tasks, we see V2T is the easiest with the highest quality and faithfulness. V2T is also the most studied task in literature. For the other three tasks, faithfulness is considerably higher than quality. Improving video and audio reconstruction in VQ-VAE can potentially lead to higher quality. This also suggests that humans can reasonably ignore generation quality in faithfulness evaluation. 
%\SZ{The R.Sim. score is influenced by both the generation model and the retrieval model. Although A2V shows slightly higher faithfulness in human evaluation than T2V, R.Sim. shows the opposite. This may be because the video-audio retrieval model is strong (as seen by high faithfulness) and so can easily detect the imperfections in the generated sample, or because it is not well calibrated on generated data (it was trained on real data). As such, unlike human evaluation, generation quality likely affects faithfulness as evaluated by retrieval models (e.g., using R.Sim).}

\section{Conclusion}
We introduce MUGEN -- a closed world, large-scale multimodal dataset based on a significantly enhanced version of the platform game CoinRun~\cite{cobbe2019quantifying}. MUGEN has videos, human-annotated text descriptions, automatically generated templated text descriptions, frame-level pixel-accurate semantic segmentation maps, as well as audio. The multiple modalities and rich annotations in MUGEN enable research progress in various tasks in multimodal understanding and generation without requiring web-scale data or massive compute. We explore retrieval and generation between every pair of modalities.
To evaluate generative models, we establish a human evaluation protocol by calibrating towards the ground-truth samples, making it easier to compare performance and show progress. \revision{The MUGEN dataset, the modified game engine, our training code and models, and the human evaluation UIs can be found at: \url{https://mugen-org.github.io/}}.

\clearpage
\bibliographystyle{splncs04}
\bibliography{egbib}
\clearpage
\appendix
\section{Dataset Collection}
\label{appendix:collection}

\subsection{Full list of modifications}

Here is a full list of the modifications we made to the game environment:
\begin{itemize}
    \item Added game audio with two layers: sound effects and background music. Background music consists of 2 songs corresponding to the space and snow themes. There are $8$ sound effects corresponding to Mugen's core actions: walk, jump, collect coin, kill monster, power-up, climb ladder, bump head, die. Each sound effect is triggered by these actions, and one sound effect plays at a time. Background music is layered with the sound effect audio to produce the full audio track.
    \item Selected a subset of assets for Mugen, the background, and ground to simplify the visual world to a single protagonist and two themes (space and snow). 
    \item Slowed game physics to reduce the pace of gameplay.
    \item Increased the camera zoom to enlarge the relative size of characters.
    \item Enabled Mugen to jump on and kill some monsters (snail, worm, face).
    \item Added animations for when Mugen dies or kills a monster.
    \item Added a gem and power-up mode for Mugen. When Mugen collects a gem, a bubble shield is placed around Mugen which protects it from being killed by any monster. The shield disappears when Mugen collects a coin.
    \item Removed the camera tracking of Mugen in the y-axis. Now, the camera only follows Mugen's movements in the x-axis. We also reduced the map height because now the camera is stable in the y-axis.
    \item Added the barnacle and frog monsters to the game.
    \item Added hopping animations for the ladybug and frog monsters.
    \item Adjusted the move speeds and abilities of monsters so none are identical in motion and ability.
\end{itemize}
Before and after videos highlighting these modifications are in the supplementary video.

\subsection{Manual Text}
\begin{figure}[h]
\includegraphics[width=12cm]{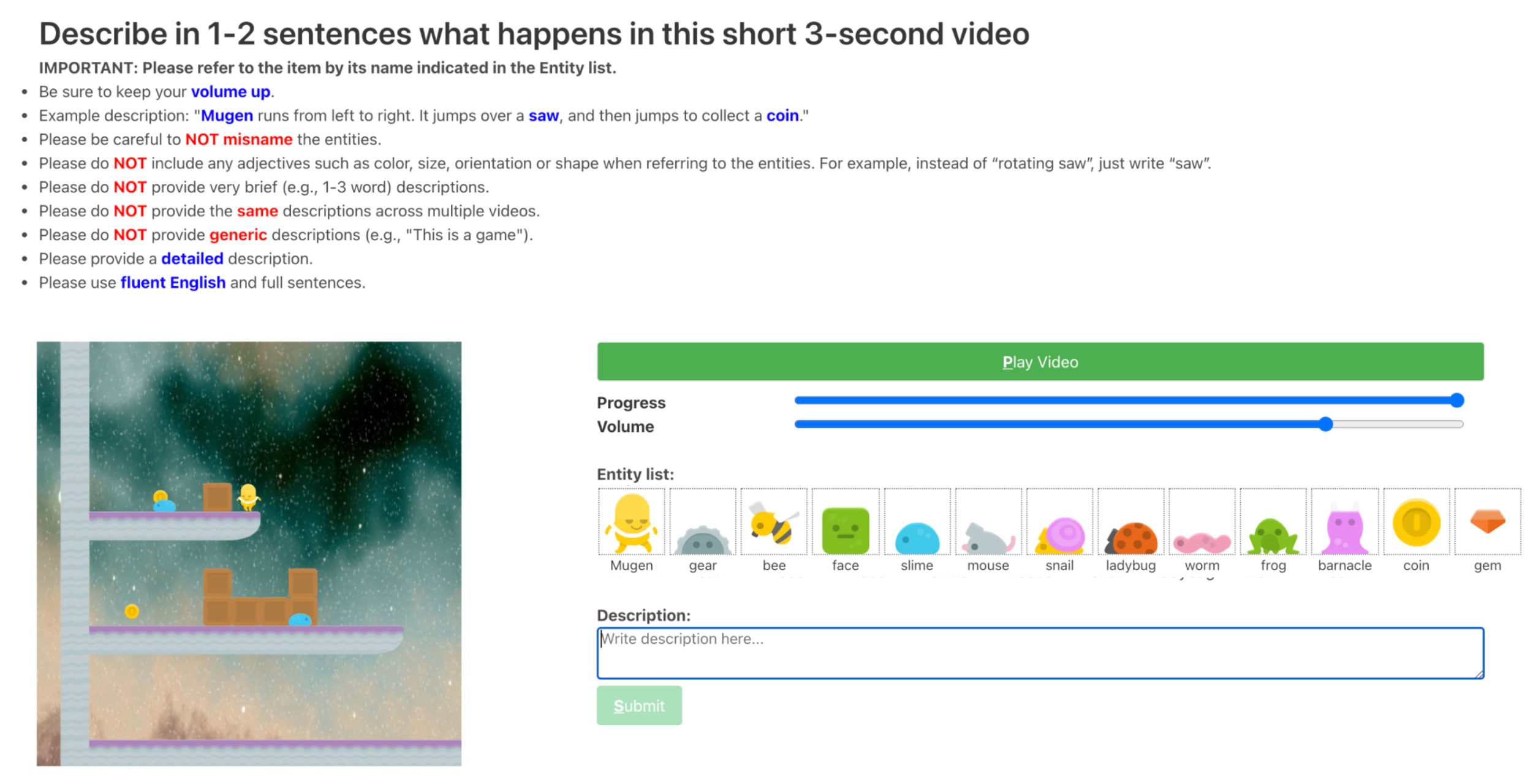}
\caption{Annotation interface to collect text descriptions for each video clip.}
\label{fig:ann}
\centering
\end{figure}

We split the recorded gameplay videos into $3.2$ second clips and collected text descriptions by asking annotators to refer to each character with their specific names. Annotators can also adjust the video playback and volume. Figure~\ref{fig:ann} shows the annotation interface. 

To ensure the annotation quality, we enforce several rules on the UI before any description can be submitted, such as: 1) Mugen must be mentioned in the description; 2) the video must be played; 3) the description length must be more than $3$ words. 
In addition, we performed several post-processing steps to remove low quality text annotations. 
Specifically, we randomly sampled collected descriptions and manually inspected annotation quality to identify annotators who write poor quality annotations. We blocked these annotators from writing more descriptions and removed all existing annotations. We also removed text descriptions that are no more than $20$ characters. In total, we filtered out $18,449$ descriptions from blocked annotators and $4,778$ short descriptions. After filtering, MUGEN consists of $378,902$ text descriptions for $375,368$ video clips, where a very small portion of the clips have more than one description.

\subsection{Auto-Text}
Since we have accurate metadata saved for all game elements and events, we can apply a template-based algorithm to automatically generate textual descriptions for each video in the following steps:

\begin{itemize}
  \item Based on Mugen's pose (e.g. jump, walk, climb) saved in metadata for each frame, we merge frames with the same pose into a segment.
  \item We further merge segments with the same pose that are only interrupted by a few frames of other poses. This is to deal with the cases like Mugen jumping repeatedly. In this process, we also count the number of segments merged (to generate a phrase later).
  \item For each segment, we generate one short phrase based on a pre-defined template. For example, for a ``jump" segment, we consider the following:
  \begin{itemize}
    \item The start and end height to decide whether Mugen jumps higher, lower, or on the same level.
    \item How many times Mugen jumped (saved from the previous merging step).
 	  \item Whether the jump has horizontal movement (left, right, or no move).
    \item Whether Mugen jumped over any enemy character.
    \item What kind of surface Mugen lands on.
    \item Whether the jump killed any enemy character in the end.
  \end{itemize}
  \item An example templated phrase generated with the above information: ``jumps up to the right over a snail to a platform". Note the character ``Mugen" is not in the phrase here; it's only added once we merge all phrases in the next step.
  \item Finally, we merge all short phrases for each segment into the full auto-text description, connect them with ``and", and put ``Mugen" at the beginning. To avoid excessively long descriptions, segments with less than 5 frames are filtered out at this step. An example of merged auto-text: ``Mugen walks to the right, and jumps up to the right to a ladder, and killed by a frog".
\end{itemize}

\begin{figure}[h]
\includegraphics[width=12cm]{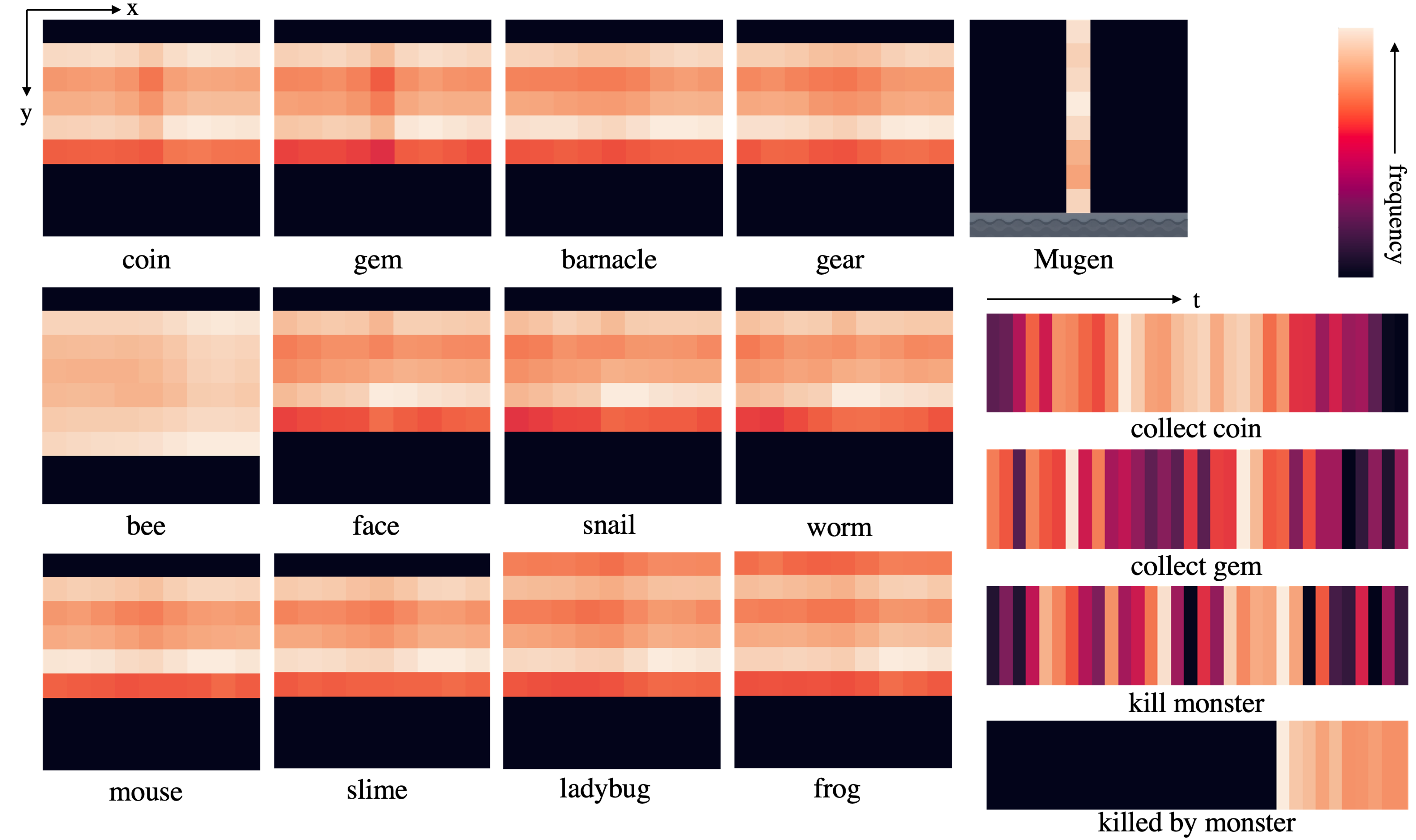}
\caption{Location heat maps for all characters and objects and temporal heat maps for the $4$ classes of interactions. All heat maps are shown on log scale to make tail events more visible.}
\label{fig:heatmap}
\centering
\end{figure}

\section{Dataset Statistics}
\label{appendix:statistics}

To better understand the distribution of entities and interactions in our dataset, we construct position and temporal heat maps. Specifically, for each character and object, we compute the distribution of positions in each frame in our $3.2$s video dataset, where the position is the center of the entity asset. For each of the $4$ classes of interactions, we compute the distribution of interaction time over the $96$ frames of each video.

Figure~\ref{fig:heatmap} shows these location and temporal heat maps. We see that all of the entities primarily lie on horizontal bands that correspond to the various $y$ positions of platforms (all entities besides Mugen and bee only appear on platforms). Mugen is always centered along the $x$-axis. Also, we observe that monsters that are able to be killed by Mugen (snail, worm, face) tend to appear closer to the center of the frame. This occurs because Mugen has incentive to approach these monsters and jump on them. Other monsters less frequently appear in the center because Mugen avoids them. Gem and coin also appear less often in the center because they disappear when Mugen collects them. Another observation is that all entities besides Mugen occur slightly more frequently on the right side of the frame than the left, which is due to map procedural generation being slightly biased towards placing platforms on the right side of the map. As for temporal heat maps of interactions, we see that all interactions are more or less uniformly distributed except ``killed by monster", which only occurs at the end of videos because the level ends once Mugen dies.

\section{Human Evaluation User Interface}

We demonstrate the user interface used in the human evaluation for multimodal generation, as shown in Figure~\ref{fig:eval_ui}. Figure~\ref{fig:eval_t_q}-\ref{fig:eval_a_q} are used for quality evaluation, while Figure~\ref{fig:eval_v2t_f}-\ref{fig:eval_v2a_f} are used for faithfulness evaluation.

\begin{figure}[h]
    \centering
    \begin{subfigure}{\textwidth}
      \includegraphics[width=12cm]{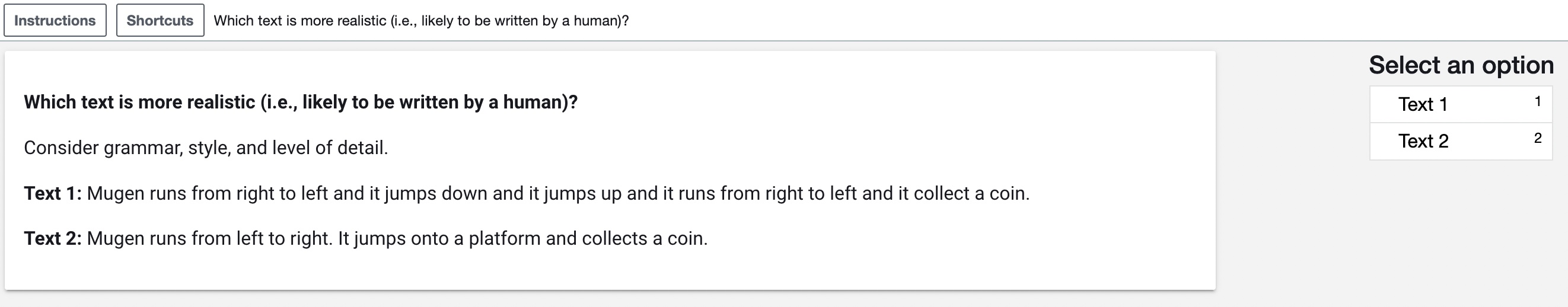}
      \caption{Text quality evaluation.}
      \label{fig:eval_t_q}
    \end{subfigure}
    \begin{subfigure}{\textwidth}
    \includegraphics[width=12cm]{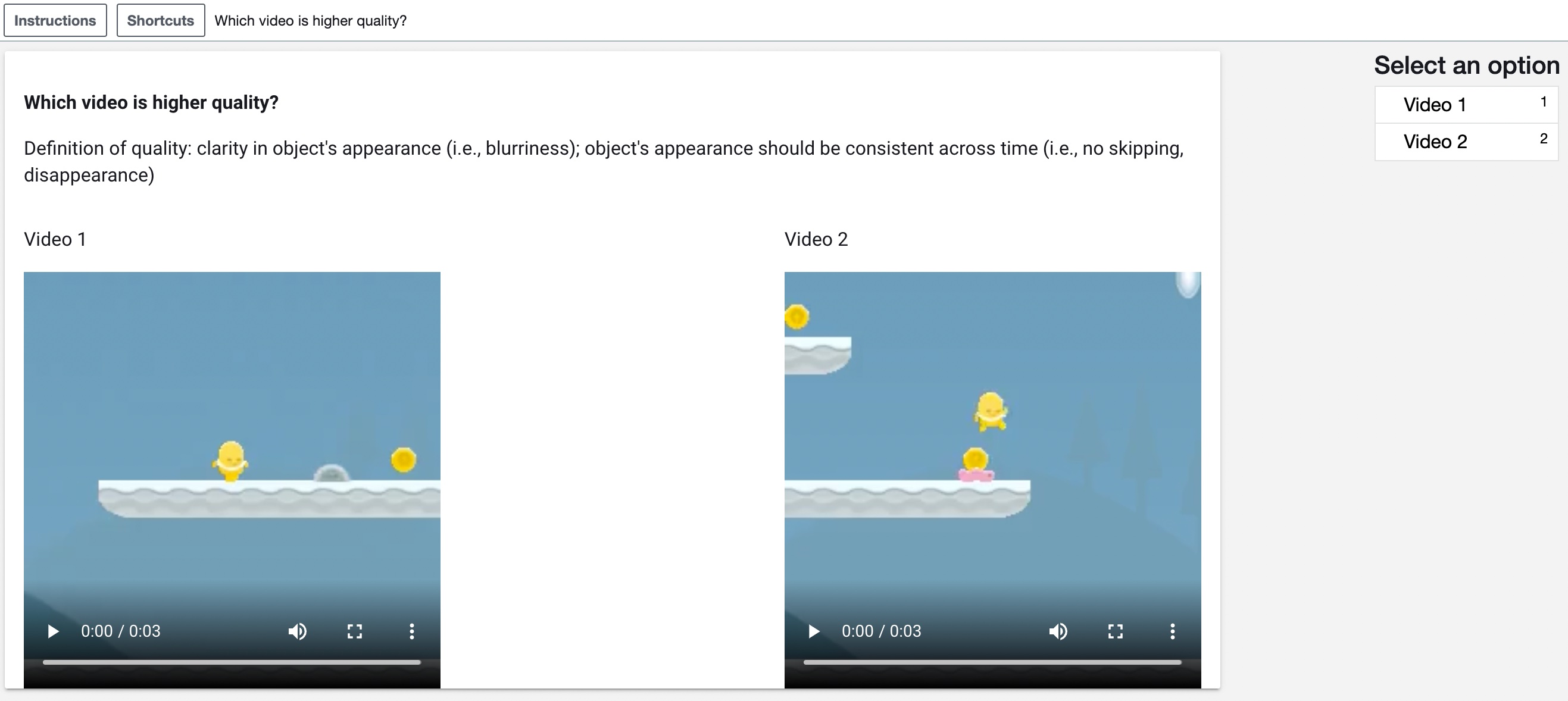}
     \caption{Video quality evaluation.}
     \label{fig:eval_v_q}
    \end{subfigure}
\end{figure}
\begin{figure}[h]\ContinuedFloat
    \begin{subfigure}{\textwidth}
    \includegraphics[width=12cm]{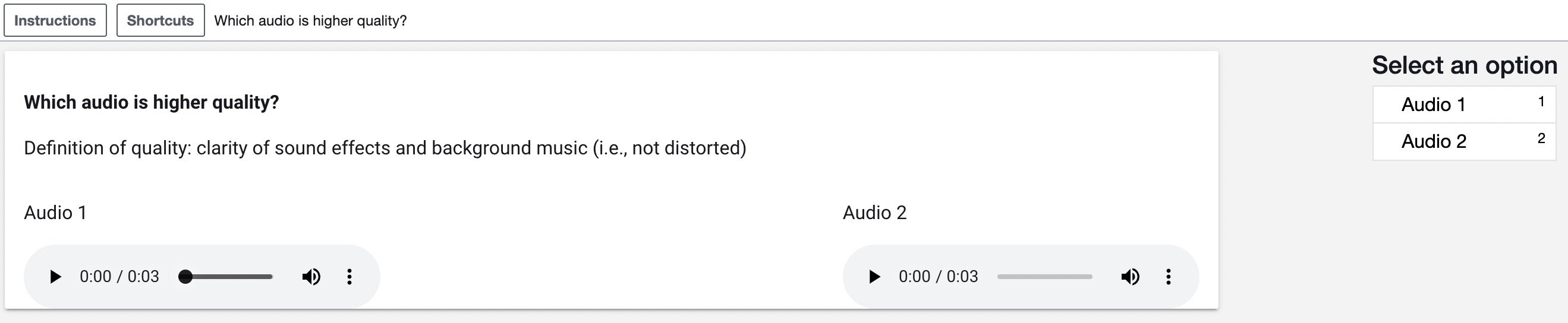}
     \caption{Audio quality evaluation.}
     \label{fig:eval_a_q}
    \end{subfigure}
    \begin{subfigure}{\textwidth}
    \includegraphics[width=12cm]{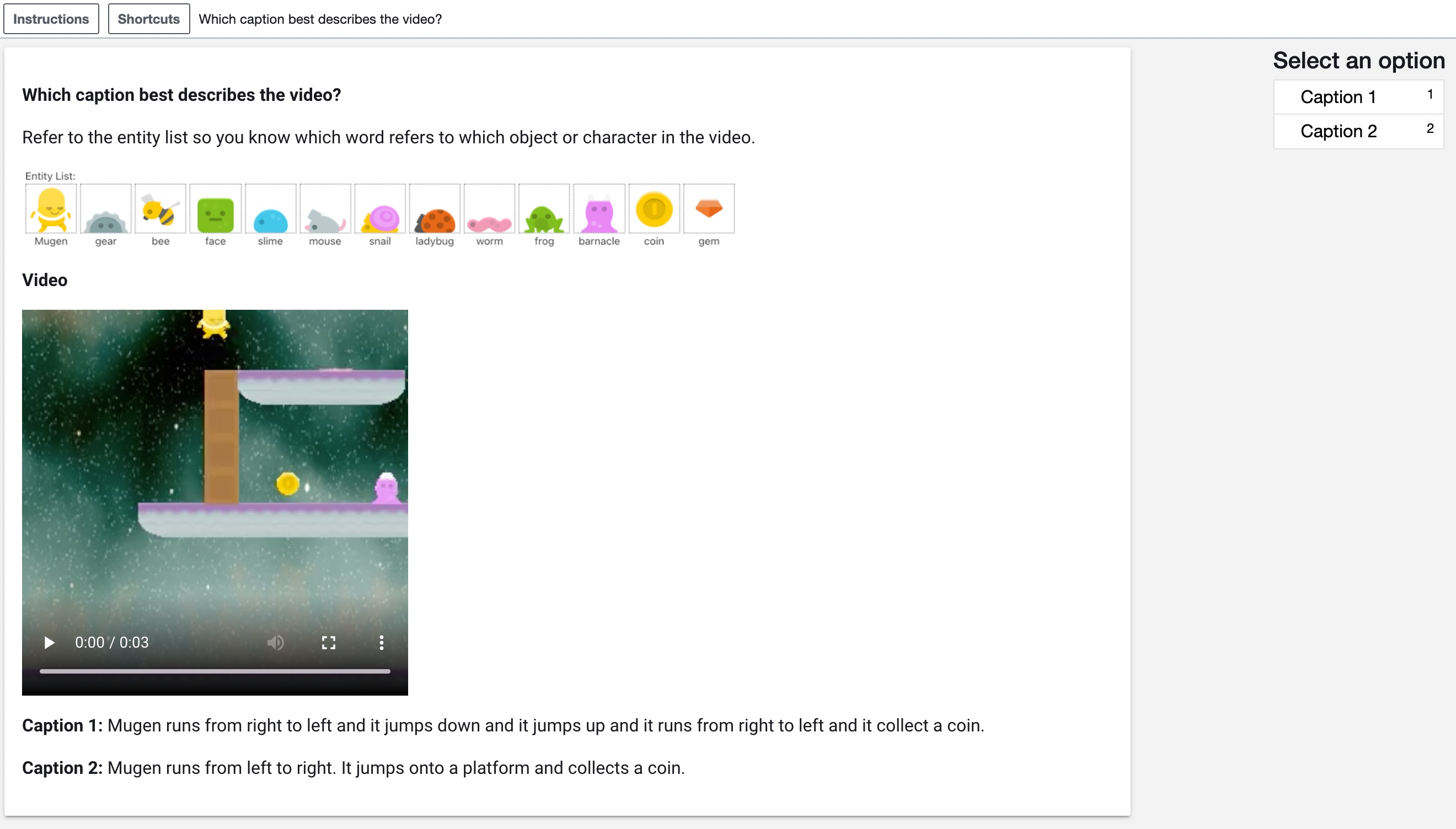}
     \caption{Video-to-text faithfulness evaluation.}
     \label{fig:eval_v2t_f}
    \end{subfigure}
    \begin{subfigure}{\textwidth}
    \includegraphics[width=12cm]{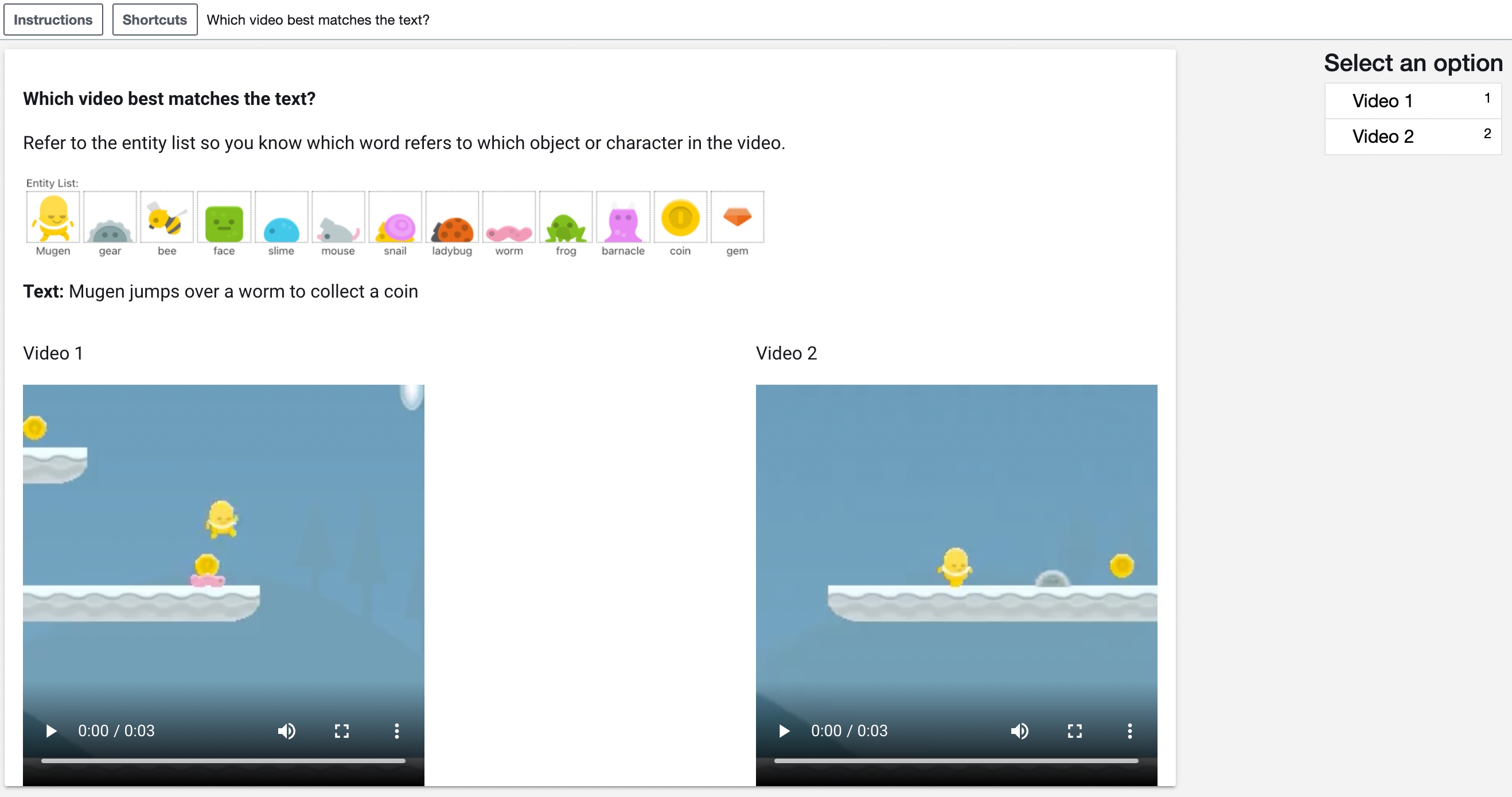}
     \caption{Text-to-video faithfulness evaluation.}
     \label{fig:eval_t2v_f}
    \end{subfigure}
\end{figure}
\begin{figure}[h]\ContinuedFloat
    \begin{subfigure}{\textwidth}
    \includegraphics[width=12cm]{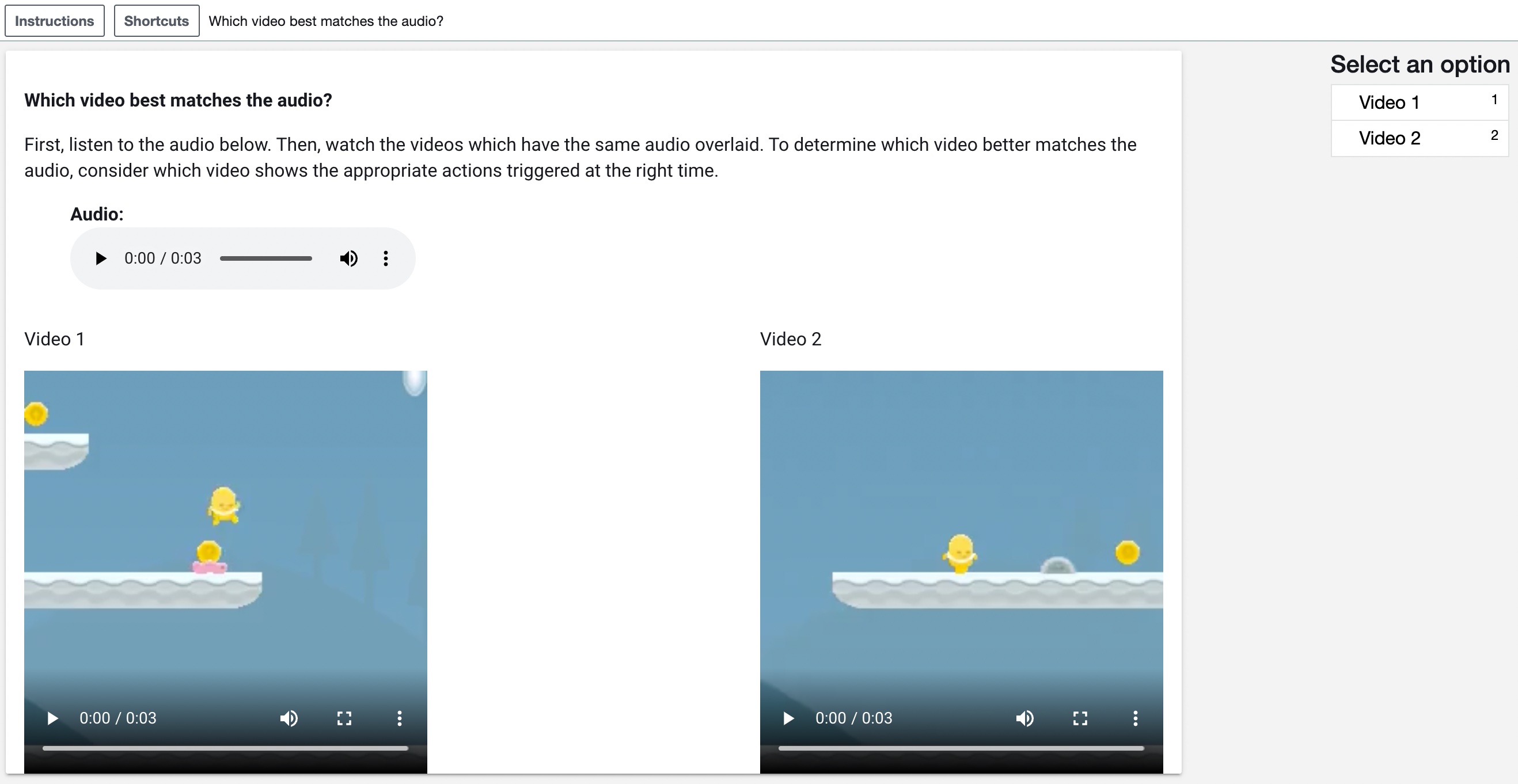}
     \caption{Audio-to-video faithfulness evaluation.}
     \label{fig:eval_a2v_f}
    \end{subfigure}
    \begin{subfigure}{\textwidth}
    \includegraphics[width=12cm]{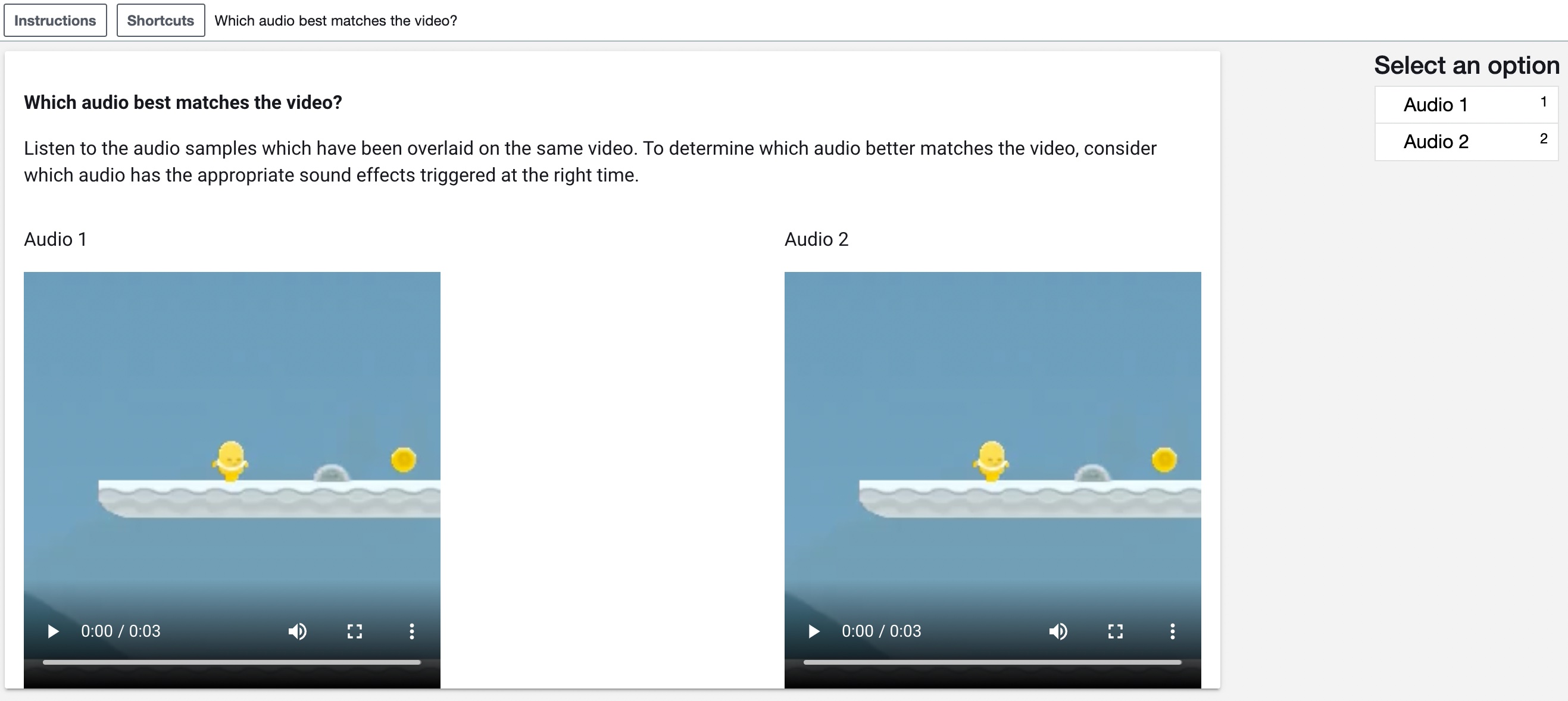}
     \caption{Video-to-audio faithfulness evaluation.}
     \label{fig:eval_v2a_f}
    \end{subfigure}
    \caption{Human evaluation user interface.}
    \label{fig:eval_ui}
\end{figure}
\end{document}